\documentclass[letterpaper, 10 pt, conference]{ieeeconf}  

\IEEEoverridecommandlockouts                              

\usepackage[utf8]{inputenc} 
\usepackage[T1]{fontenc}    
\usepackage{hyperref}       
\usepackage{url}            
\usepackage{booktabs}       
\usepackage{amsfonts}       
\usepackage{nicefrac}       
\usepackage{microtype}      
\usepackage{xcolor}         
\usepackage{amsmath}        
\usepackage[linesnumbered,ruled]{algorithm2e}
\usepackage{graphicx}
\usepackage{subfigure}
\usepackage{algpseudocode}
\usepackage{caption}

\title{\LARGE \bf
NeuroLoc: Encoding Navigation Cells for 6-DOF Camera Localization 
}

\author{
Xun Li$^{1}$, Jian Yang$^{2}$,  Fenli Jia$^{2}$, Muyu Wang$^{1}$,  
Jun Wu$^{1}$, Jinpeng Mi$^{3}$, Jilin Hu$^{1}$, \\ Peidong Liang$^{4}$,  Xuan Tang$^{1}$, Ke Li$^{2}$, Xiong You$^{2}$, Xian Wei$^{1\dagger}$
\thanks{
$^{1}$Software Engineering Institute, East China Normal University; $^{2}$School of Geospatial Information, Information Engineering University; $^{3}$University of Shanghai for Science and Technology; $^{4}$Fujian (Quanzhou) Institute of Advanced Manufacturing Technology, China; $^{\dagger}$Corresponding Author.
    }%
}

\begin{document}

\maketitle
\thispagestyle{empty}
\pagestyle{empty}

\begin{abstract}

Recently, camera localization has been widely adopted in autonomous robotic navigation due to its efficiency and convenience. However, autonomous navigation in unknown environments often suffers from scene ambiguity, environmental disturbances, and dynamic object transformation in camera localization. 
To address this problem, inspired by the biological brain navigation mechanism (such as grid cells, place cells, and head direction cells), we propose a novel neurobiological camera location method, namely NeuroLoc. Firstly, we designed a Hebbian learning module driven by place cells to save and replay historical information, aiming to restore the details of historical representations and solve the issue of scene fuzziness. 
Secondly, we utilized the head direction cell-inspired internal direction learning as multi-head attention embedding to help restore the true orientation in similar scenes. 
Finally, we added a 3D grid center prediction in the pose regression module to reduce the final wrong prediction. 
We evaluate the proposed NeuroLoc on commonly used benchmark indoor and outdoor datasets. 
The experimental results show that our NeuroLoc can enhance the robustness in complex environments and improve the performance of pose regression by using only a single image.

\end{abstract}

\section{INTRODUCTION}

%
%
%
%
Camera localization is one of the most essential tasks in machine vision. It aims to determine the camera's position and orientation by analyzing the scene's visual information without relying on external data. 
At present, it has been widely employed in autonomous driving~\cite{9411961}, robotic navigation~\cite{SUGIHARA1988112}, and augmented reality~\cite{4637573}.

The classic camera pose estimation problem can be solved by a matching algorithm based on structural features~\cite{Li_2020_CVPR, Sarlin_2019_CVPR} or image retrieval algorithms from large-scale database~\cite{sattler2019understanding, WANG2024109914}. 
%
However, they often require a large amount of storage space to store maps and are highly sensitive to changes in lighting and object occlusion in outdoor scenes. 
%
Various deep learning algorithms for camera localization have been undertaken because of their low cost and high efficiency. 
For example, Posenet~\cite{Kendall_2015_ICCV} can directly predict the global pose from a single image without other manual constraints. 
Its variants utilize different feature extraction networks~\cite{Melekhov_2017_ICCV} and geometric constraints~\cite{ QIAO202311, Kendall_2017_CVPR} to enhance the performance. 
However, in the outdoor scene, there are a lot of dynamic objects and light changes, which leads to the lack of robustness of the network as a whole. Recent work has considered using multi-view input to learn scene features~\cite{ Brahmbhatt_2018_CVPR} or using continuous frame images to learn temporal and spatial context features~\cite{ 2017Image}.

%
To explore the more robust solution to outdoor camera localization, researchers find that animals in nature exhibit excellent self-positioning abilities~\cite{mouritsen2018long} in complex wilderness environments and can perform accurate long-distance navigation.
Specifically, biologists have found that place cells~\cite{o1971hippocampus}, grid cells~\cite{hafting2005microstructure}, and head direction cells~\cite{taube1990head} in the brain support this positioning ability. 
Specifically, head direction cells always provide compass information for animal orientation.
Grid cells combine directional and velocity signals to provide a metric for determining position during localization~\cite{gao2021path}, generating an activity pattern that facilitates navigation. 
The place cell will store the localization information in the scene to form a spatial storage and activate it when revisiting the location.
We argue that this kind of navigation ability of animals can inspire solving the robustness problem in visual localization.
%

\begin{figure*}
 \centering
    \includegraphics[width=17cm]{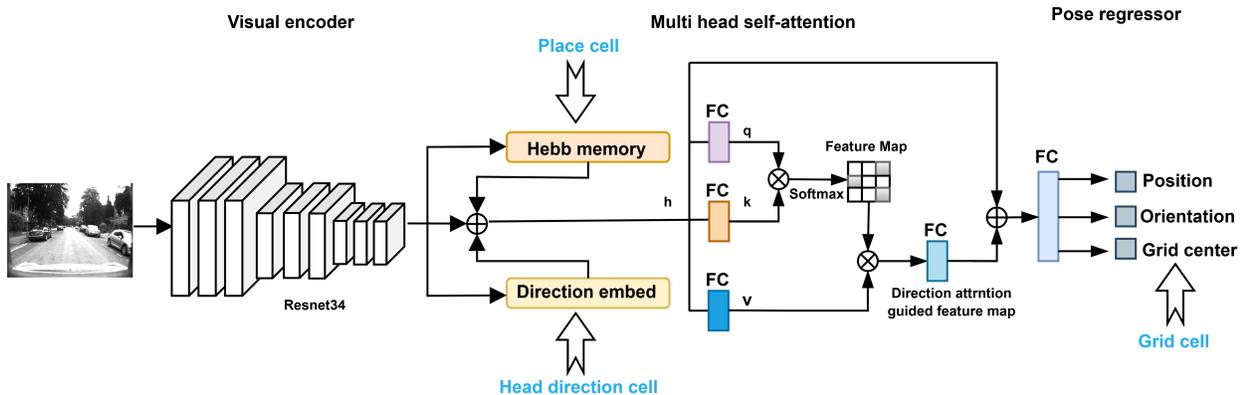}
    \caption{\textbf{An overview of the proposed NeuroLoc framework.} It includes a visual encoder (extracting scene features from a single image), a Hebbian Storage Module (storing and reading scene information), a pose regression module (directional attention is used to map attention features to camera poses, and a 3D grid module is used to predict grid center positions).
    \label{Main_framework}
}	
\vspace{-6mm}
\end{figure*}

In this work, we proposed the NeuroLoc model inspired by the biological mechanisms of navigation cells. 
%
We address the challenges of scene ambiguity, environmental disturbances, and dynamic object transformations in camera localization by incorporating the role of grid cells, place cells, and head direction cells.
The pipeline of NeuroLoc could be grouped into the following three aspects (see Figure~\ref{Main_framework}):
 \textbf{3D grid position prediction:} Drawing on the grid-like discharge encoding of grid cells, we propose a 3D grid regression network to predict the center position of the grid. 
\textbf{Place cell encoding:} We designed a Hebbian learning rule-constrained historical information access network inspired by the access to localization information of place cells.
\textbf{Head direction cell encoding:} The activation area of head direction cells follows the direction change of the animal's head. We have designed an attention network that integrates directional distribution. 
The overall network predicts absolute pose and 3D grid center position from a single image, achieving state-of-the-art performance across indoor and outdoor scenarios.

The main contributions of this work are as follows:
\begin{itemize}

\item A Hebbian learning rule-constrained place cell module is proposed for storing and reading historical information, which helps refine image features and reduce scene ambiguity.
\item We present a pose regression module integrating a directional attention mechanism and grid position constraints to learn the relationship between geometric features and real positions. This helps to solve dynamic object transformation problems and reduce overall errors.
\item By visualizing the feature saliency map of attention, we have demonstrated that directional distribution design can learn stable geometric features.

\end{itemize}

\section{Related Work}

\subsection{Deep Learning Methods in Camera Localization}

%
Recently, the method based on deep learning has achieved good performance in camera positioning tasks.
The pioneering algorithm Posenet~\cite{Kendall_2015_ICCV} and some of its variants~\cite{Melekhov_2017_ICCV, cai2019hybrid} use deep neural networks (DNNs) to learn camera pose from a single image directly. These algorithms are time efficient, but they lack robustness in some complex scenes, such as scenes with textureless areas, local similarity, and light changes. For this kind of problem, people propose a variety of solutions are proposed from multiple perspectives, such as spatial continuity, geometric features, and data enhancement. 
PoseNet+LSTM~\cite{2017Image} uses LSTM units on CNN. This method uses image continuity in time and space to obtain more structural features and improve positioning performance. MapNet~\cite{Brahmbhatt_2018_CVPR} introduced additional information from IMU, GPS, and visual SLAM systems as constraints to ensure pose consistency between consecutive frames. Another method is to use geometric constraints of paired images~\cite{ QIAO202311, Kendall_2017_CVPR} and synthesize new training data~\cite{Huang_2019_ICCV, SARIGUL2023110} or introduce neural mapping pose map optimization of models~\cite{Parisotto_2018_CVPR_Workshops}. In this work, we adopt a new strategy for designing a DNN model inspired by navigation cells encoding for network self-regulation. This method can automatically learn geometric robust features that contribute to pose regression and store them persistently.

\subsection{Navigation Cells Inspired Localization}
%
Recently, the methods inspired by animal navigation cells have completed some spatial positioning~\cite{zeng2020neurobayesslam, yu2023nidaloc, zhou2017brain} and path-planning tasks~\cite{chen2019brain} in the field of robot navigation. 
Ratslam~\cite{1307183} was the first to use a computational model of rodent hippocampus to perform vision-based SLAM, mapping movement state information to the activity state changes of pose cells based on a competitive attractor network and combining visual input to achieve localization function. 
NeuroSLAM~\cite{2019NeuroSLAM} constructs a joint pose cell module to represent 4DoF poses and incorporates the input of visual odometry to achieve the composition and updating of multi-layer empirical maps. 
However, most navigation cell-inspired methods require external visual input and self-movement cues\cite{yu2023brain,liao2022brain}. 
We propose a brain-inspired localization method requiring only a single image input to predict 6-Dof poses.
By designing an internal module and attention module with cell-like functionality based on the traditional APR (Absolute camera Pose Regression) architecture, the network can automatically learn and store geometric features related to localization, which helps alleviate the problem of scene ambiguity and improve the overall localization performance.
\section{The Proposed NeuroLoc}
This section introduces the details of the proposed NeuroLoc.
Figure~\ref{Main_framework} shows the overall framework of the proposed NeuroLoc, which mainly learns 6-Dof camera pose and 3D grid center position from a single image. The proposed network consists of three components: a Visual Encoder, a Hebbian Storage module, and a Pose Regression module.
%
\subsection{Visual Encoder}
%
The first step in NeuroLoc is to learn implicit features from the initial image using a visual encoder. Previous work has shown the excellent performance of CNNs in camera pose regression.
Considering the compatibility with subsequent network modules and the stability of the overall network, we used a CNN-based network, ResNet34~\cite{He_2016_CVPR}, as the visual encoder of the network. 

To better adapt the visual encoder to our network module, we use a full 2048-dimensional connection to replace the original 1000-dimensional fully connected layer and remove the softmax layer for classification. 
%

In the following, 
the proposed NeuroLoc enhances the localization of agent by using Hebbian plasticity rules of place cells in Section~\ref{Sec_Hebb}, and improves the pose estimation by using head direction cells and grid cells in Section~\ref{Sec_poseReg}.
\subsection{Hebbian Storage Module}
\label{Sec_Hebb}
%
%
As shown in~\cite{leutgeb2005independent}, place cells can encode and store historical scenes and activate them when returning to a specific location in the historical scene. 
Inspired by this mechanism of place cells, we propose a Hebbian storage module to save and refine the features of historical scenes.
This will help address the issue of scene ambiguity caused by the large areas of scenes, which is a challenge for recent camera pose prediction models.
We hope that we can improve the fuzziness of the scene by using the ability of place cells to continuously encode and update scene features in the temporal dimension.
Because the mammalian brain uses Hebbian synapses~\cite{citri2008synaptic}, we consider using Hebbian plasticity rules to update scene features. 
Hebb plasticity rules reveal that the strength of connections between neurons varies with the activities of presynaptic and postsynaptic neurons. 
\begin{figure}[htp]    
    \centering
    \includegraphics[width=8cm]{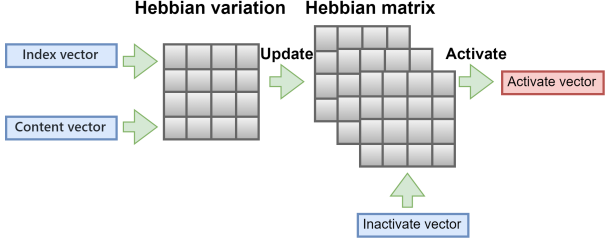}
    \caption{\textbf{Overview of the Hebbian storage module.} The input features will be expanded into index vectors, context vectors, and inactivate vectors, and then the storage matrix will be updated using Hebbian rules (persistent storage of scene features). The inactive vector is multiplied by the Hebbian matrix to obtain the activated vector.}

\end{figure}
%

\textbf{Updating of Hebbian-based Rules}:
We have modified the update formula of Hebbian-based Rules based on Dual OR~\cite{vasilkoski2011review} to restrict the network from learning scene features according to synaptic plasticity-like rules. Hebbian-based rules expression is defined as: 
\begin{equation}
          W_{i} = \eta_{i}  (k \bullet v - k \bullet W_{i-1}) 
          ,W = Concat(W_{i})
\end{equation}
Firstly, the global features output by the visual encoder will be expanded to $k \in R^{B\times 2048\times 1}$ and $v \in R^{B\times 1\times 2048}$. ($B$ is the batch size). 
Here, $\eta_{i} k v $ is the Hebbian correlation term, $\eta_{i} k W_{i-1}$ is the penalty term, $W_{i}$ is the current storage matrix, $W_{i-1}$ is the past storage matrix, and $\eta_{i}$ is a dynamic attenuation parameter.


%
\textbf{Activating of Hebbian-based Rules}:
We use matrix multiplication to extract activation vectors $q$ from the storage matrix $W \in R^{B\times 2048\times 2048}$, and transform them through fully connected layers, following the normalization, and RELU functions. 
Finally, we use a residual module to convert the activation vector into a positional encoding $x_{pc}\in R ^ {B \times 2048} $. 



\subsection{Pose Regression Module}
\label{Sec_poseReg}
%
%
Inspired by head direction cells and grid cells, we propose a novel pose regression module for the APR framework. The main process of the pose regression module is as follows: the Biologically Plausible Direction Attention Module learns geometric features that help with localization in dynamic object changes and sent to the fully connected layer and 3D grid center module with global features to predict the absolute camera pose and grid center position.

\subsubsection{Biologically Plausible Direction Attention Module}

%
Recent camera positioning models have encountered issues with abnormal rotation predictions due to dynamic objects.
Recent research found that each head direction cell in the brain learns a directional preference, which is sensitive to the rotations~\cite{WANG2024109914}.
%
%
When the angle between the true head direction and the directional preference in the head direction cell is smaller, the cell discharge becomes stronger~\cite{WANG2024109914}, and the discharge frequency reaches its minimum at around 45°. 
Therefore, inspired by the biological mechanism of head direction cells, we propose an internally constrained multi-head attention module for learning the relationship between feature space and true camera orientation. 
Specifically, we add a position encoding to the input of the attention network to imitate the direction preference mechanism of head direction cells and use the attention network to learn the mapping relationship between internal features and camera orientation.
%
%
This helps to learn meaningful directional expressions from dynamic objects.

\begin{figure}[htp]    
    \centering
    \includegraphics[width=8cm]{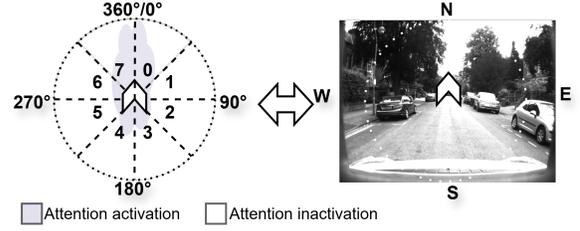}
    \caption{\textbf{The left image shows the activation status inside the feature after embedding direction encoding.} The image on the right shows the true direction in the real world.}
    \label{hd_cell}
\end{figure}


%
Figure~\ref{hd_cell} depicts the working mechanism of the proposed biologically plausible attention module. 
%
In the direction constraint in Figure~\ref{hd_cell}, we divide the circle of the plane into $d\in R^{8\times 2\pi/8}$ regions, which is used to simulate the response region of the head direction cell.
To activate a specific interval, we created a weight parameter $\xi\in R^{8\times 256}$ with a learnable range limited to $[0,1]$. 
It is used to balance the activation weights of each interval and simulate the activation state of specific head direction cells in a head direction cell population. 
The direction values of each interval obtained in the end will be encoded through trigonometric transformation.
\begin{equation}
          x_{hd} = x_{pc} + \xi \ast \sin{d/2}.
\end{equation}
%


To compute the attention map, we transform $x_{hd}$ to the linear space of $\theta(x_{hd})$ and $\psi(x_{hd})$, and calculate the similarity of scaling dot product of $x_{hd}$ in $\theta$ and $\psi$ as 
%
\begin{equation}
          S(x_{hd}) =\frac{\theta(x_{hd})\psi(x_{hd})}{\sqrt{D_k}},
\end{equation}
where $D_k$ is the dimension of input $x_{hd}$,  $\theta(x_{hd})=W_{\theta}x_{hd}$, $\psi(x_{hd})=W_{\psi}x_{hd}$. 
We use softmax to normalize $S(x_{hd})$ and multiply $g(x_{hd})=W_{g}x_{hd}$ to get the attention vector $h_{i}$ of a single head is as follows:
\begin{equation}
            h_{i} = Softmax (S(x_{hd})) g(x_{hd}),
\end{equation}
and then splice it to get the final multi-head attention vector $H=Concact (hi)W^O,i=1,2$. $W^O$ is a learnable parameter matrix.
At the same time, we add the residual structure to the output of multi-head attention. Finally, our mathematical expression is as follows:
\begin{equation}
         y = Softmax (x^{T}W_{\theta}^{T}W_{\psi}x) W_{g}x + x.
\end{equation}

\subsubsection{3D Grid Module} 
\begin{figure}[htp]    
    \centering
    \includegraphics[width=8cm]{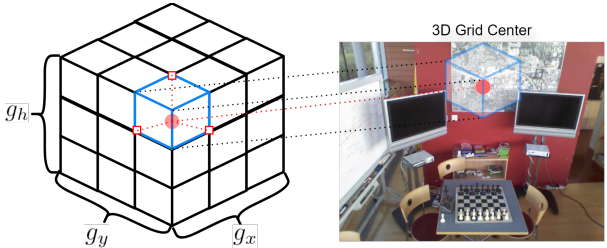}
    \caption{The left figure shows that we constructed $n$ 3D grids in 3D space, where $g_x$, $g_y$, and $g_h$ represent the grid boundaries (map boundaries) on the three coordinate axes, and the red dots represent the center positions of the 3D grids. The figure on the right shows that the center of our 3D mesh is directly calculated in the real world.}
    \label{grid_cell}
    \vspace{-2mm}
\end{figure}
Inspired by the biological mechanisms of grid cells, we have added a 3D grid center prediction to the standard prediction of the pose regression module. 
According to our investigation, grid cells are neurons that regularly fire as an animal moves through space, creating a pattern of activity that aids navigation. To simulate the mechanism of grid cells, we perform equidistant 3D grid partitioning in a real scene and predict the center position of the grid in the real scene for each image, as shown in Figure~\ref{grid_cell}. In the pose regression module, we created a fully connected layer for grid center prediction, similar to predicting the corresponding grid cell block of each image in the scene by grid position prediction, which shares the weights of the main network architecture with position prediction and rotation prediction.

\subsection{Loss Function}
%

In this work, the proposed network outputs the absolute position (a translation vector $t\in R^{3}$ and a quaternion-based rotation vector $q \in R^{4}$) and the center position of the grid (a position vector $g\in R^{3}$). The parameters of the neural network are optimized by minimizing the following loss function $L$:
\begin{equation}
\begin{split}
         l_{pose} &= \lVert p-\overline{p} \lVert_{1}e^{-\alpha}+\alpha +\lVert \log q-\log \overline{q} \lVert_{1}e^{-\beta}+\beta ,\\
       L &= l_{pose} + \lVert g-\overline{g} \lVert_{1}e^{-\gamma}+\gamma,
       \end{split}
\end{equation}
%
where $l_{pose}$ is the loss for measuring the offsets of position and direction,
 the hyperparameter factors  $\alpha$, $\beta$, and $\gamma$ control the influence of the position loss, rotation loss, and grid loss on the final solution,
$p$, $q$, and $g$ represent the predicted absolute position, direction, and grid center position, $\overline{p}$, $\overline{q}$ and $\overline{g}$ represent the real absolute position, direction, and grid center position. Therein, $\log q$ is the logarithmic form of the unit quaternion.







\section{Experiments}

\subsection{Datasets}

\subsubsection{7 Scenes} The 7 Scenes~\cite{Shotton_2013_CVPR} dataset, an indoor scene one with RGB-D images, real camera pose, and 3D models of seven rooms, has about 125 m² indoor environment. Each scene has 2-7 image sequences (500 or 1000 images per sequence) for training/testing. Its images cover textureless surfaces, motion blur, and repetitive structures, being a popular visual localization dataset.

\subsubsection{Oxford RobotCar} Oxford Robotbar dataset~\cite{doi:10.1177/0278364916679498} has 100 times of repeated driving data of a autonomous Nissan LEAF car in downtown Oxford within a year. It contains various weather, traffic, and dynamic objects, making it challenging for vision based localization tasks.

\subsection{Implementation Details}
The input of our APR model is a monocular RGB image, and the short edge of the image is scaled to 256 pixels. 
Resnet34~\cite{He_2016_CVPR} in our network is initialized using the pre-training model on the ImageNet dataset, and the rest of the components are initialized using random initialization. 
We use random color jitter for data enhancement on the Oxford Robotbar dataset by Atloc~\cite{wang2020atloc}. 
The values of brightness, contrast, and saturation are set to 0.7, and the hue value is set to $0.5$. We use PyTorch to implement our method, using Adam optimizer~\cite{kingma2014adam} and an initial learning rate of $3 \times 10^{-5}$. 
The network is trained on NVIDIA 2080Ti using the following hyperparameters: the batch size is $128$, the training batch is $1200$, the dropout rate probability of 0.5, the loss weight initialization is $\alpha=0.0$, the loss weight initialization is $\beta=-3.0$, the loss weight initialization is $\gamma=0.0$, the weight attenuation rate is $5 \times 10^{-3}$, and the number of grids is 40. 
\subsection{Experiments on the 7 Scenes Dataset}
\begin{table*}[htp]

    \centering
\caption{\textbf{Localization results for the 7 Scenes dataset (indoor localization).} We report the median position/orientation error in meters/degrees for each method. The best results are highlighted in bold.} 
\label{7Scenes_result} 

\begin{tabular}{ccccccccc} 
\toprule 
Method	&	Chess	&Fire	&Heads	&Office&	Pumpkin	&Kitchen	&Stairs&	Avg \\
\midrule

PoseNet~\cite{Kendall_2015_ICCV}	&0.32/8.12&	0.47/14.4	&0.29/12.0&	0.48/7.68&	0.47/8.42&	0.59/8.64&	0.47/13.8&	0.45/9.94\\



GPoseNet~\cite{cai2019hybrid} 	&0.20/7.11&	0.38/12.3&	0.21/13.8&	0.28/8.83	&0.37/6.94&	0.35/8.15	&0.37/12.5	&0.31/9.95\\

AtLoc~\cite{wang2020atloc}	&\textbf{0.10}/4.07&	0.25/11.4	&0.16/\textbf{11.8}&	\textbf{0.17}/5.34&	0.21/\textbf{4.37}	&0.23/5.42&	\textbf{0.26/10.50}	&0.20/\textbf{7.56}\\

MLFBPPose~\cite{wang2019discriminative} 	&0.12/5.82&	0.26/11.99&	\textbf{0.14}/13.54&	0.18/8.24	&0.21/7.05&	0.22/8.14	&0.38/10.26&	0.22/9.29\\

 ViPR~\cite{Ott_2020_CVPR_Workshops}&	0.22/7.89	&0.38/12.74&	0.21/16.41&	0.35/9.59&	0.37/8.45&	0.40/9.32	&0.31/12.65	&0.32/11.01\\

NeuralR-Pose~\cite{Zhu_2021_CVPR}& 0.12/4.83 &0.27 /\textbf{8.91}& 0.16 /12.84 &0.19 /6.64 &0.22/5.45& 0.24/6.10 &0.29/10.70 &0.21 /7.92\\

IRPNet~\cite{9412225} 	&0.13/5.64&	0.25/9.67	&0.15/13.1&	0.24/6.33	&0.22/5.78&	0.30/7.29&	0.34/11.6	&0.23/8.49\\

ORGPoseNet~\cite{QIAO202311}&	\textbf{0.10/3.25}	&0.33/11.02	&0.15/13.34	&0.19/5.91&	0.20/5.42	&0.24/5.71&	0.27/10.63	&0.21/7.90\\

TransBoNet~\cite{SONG2024109975}	&0.11/4.48	&0.25/12.46	&0.18/14.00	&0.20/\textbf{5.08}&	\textbf{0.19}/4.77	&0.17/5.35&	0.30/13.04&	0.20/8.45\\

NeuroLoc(Our)	&0.12/4.37	&\textbf{0.24}/12.07
	&0.16/12.66&	0.19/6.36	&\textbf{0.19}/4.62&	\textbf{0.15/5.26}	&0.27/11.68	&\textbf{0.18}/8.14\\
\bottomrule 
\end{tabular}\vspace{-2mm}
\end{table*}

\begin{table*}[!htbp]
\centering
\caption{\textbf{Localization results of the LOOP trajectories on the Oxford Robotcar dataset (outdoor localization).} }
\label{Oxford_result} 
\begin{tabular}{c|cccccccc}
\hline
\multicolumn{1}{c|}{{Dataset} }& 

\multicolumn{2}{c}{PoseNet+~\cite{Kendall_2017_CVPR}}& \multicolumn{2}{c}{MapNet [15]}& 
\multicolumn{2}{c}{AtLoc~\cite{wang2020atloc}}& \multicolumn{2}{c}{NeuroLoc(Ours)}\\
\hline
{-}& 
{Median}& \multicolumn{1}{c|}{Mean}& 
{Median}& \multicolumn{1}{c|}{Mean}& 
{Median}& \multicolumn{1}{c|}{Mean}& 
{Median}& {Mean}\\
\hline

LOOP1&
6.88m, 2.06°& \multicolumn{1}{c|}{25.29m, 17.45°}&
5.79m, \textbf{1.54°}& \multicolumn{1}{c|}{8.76m, 3.46°}&
5.68m, 2.23°& \multicolumn{1}{c|}{8.61m, 4.58°}&
 \textbf{4.46m}, 1.72°& \textbf{8.54m}, \textbf{3.23°}\\

\hline
LOOP2&
5.80m, 2.05° & \multicolumn{1}{c|}{28.81m, 19.62°}&
4.91m, \textbf{1.67°}& \multicolumn{1}{c|}{9.84m, 3.96°}&
5.05m, 2.01°& \multicolumn{1}{c|}{8.86m, 4.67°}&
\textbf{4.69m}, 2.19°&\textbf{8.57m}, \textbf{3.94°}\\

\hline
Average&
6.34m, 2.05° & \multicolumn{1}{c|}{27.05m, 18.53°}&
5.35m, \textbf{1.60°}& \multicolumn{1}{c|}{9.30m, 3.71°}&
5.36m, 2.12°& \multicolumn{1}{c|}{8.73m, 4.62°}&
\textbf{4.57m}, 1.82°&\textbf{8.55m}, \textbf{3.58°}\\

\hline
\end{tabular}
\vspace{-2mm}
\end{table*}

\subsubsection{Results Analysis} Table~\ref{7Scenes_result} summarizes the performance of all methods. Our method achieves the best performance in all single image-based methods. Compared with the optimal model based on a single image, the positioning accuracy is improved by 10\%. In particular, NeruoLoc performs best in large textureless regions (such as pumpkin and fire). In highly textured repeated regions (stairs), the position error is reduced from 0.17m to 0.15m, and the rotation error is reduced from 5.35° to 5.26°. In other cases, NeuroLoc can also achieve accuracy similar to the benchmark. 

\subsubsection{Visualization Analysis}


Our model is superior to other models in fire, pumpkin (weak texture), and kitchen (specular reflection), which we have analyzed. In fire and pumpkin scenes, the input model from a single image will be severely affected by feature interference caused by textureless areas and highly repetitive textures. The directional attention module in our model can help the network resist meaningless feature regions, focus on meaningful scene geometric boundaries, and enhance localization robustness.
%
%

%
%

\subsection{Experiments on the RobotCar Dataset}

\begin{figure}[t]
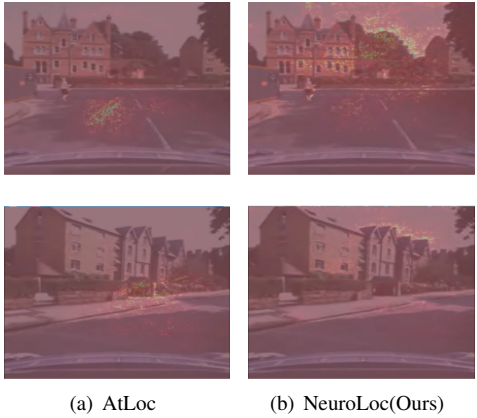

\centering  

\subfigure[AtLoc]{
\label{saliencyA}
\includegraphics[width=3cm,height=5.0cm]{img/heatmap_atloc.pdf}}
\subfigure[NeuroLoc(Ours)]{
\label{saliencyB}
\includegraphics[width=3cm,height=5.0cm]{img/heatmap_neuroloc.pdf}}
\caption{\textbf{Saliency maps of two scenes selected from Oxford RobotCar for straight driving (up) and turning (down).}}
\label{saliency}
\vspace{-6mm}
\end{figure}

\subsubsection{Result Analysis} The Oxford RobotCar dataset has the characteristics of a long collection cycle and a large area, which is very challenging for the camera localization model. Table~\ref{Oxford_result} compares our method with PoseNet+, MapNet, and AtLoc. Compared with Posenet+, the average position accuracy of LOOP1 is improved from $25.29m$ to $8.54m$, and LOOP2 is improved from $28.81m$ to $8.57m$. The overall average accuracy is $68.4\%$ and $80.7\%$ higher than PoseNet+. Compared with sequence-based MapNet, our model has significantly improved accuracy in all scenarios. Compared with AtLoc, which also contains the attention module, our model improves $3.1\%$ and $22.6\%$ in the overall average accuracy.
\begin{table}[!htbp]
\centering
    \caption{\textbf{Ablation study of NeuroLoc on Oxford RobotCar.} We report the mean position/orientation
error in meters/degrees for each method.} 
    \label{ablation}
\scalebox{1.0}{
\begin{tabular}{c|ccc}

\hline
{Dataset }& 
\multicolumn{1}{c|}{NeuroLoc-Base}&
\multicolumn{1}{c|}{NeuroLoc-Hebbian}& 
\multicolumn{1}{c}{NeuroLoc}\\
\hline
LOOP1&
\multicolumn{1}{c|}{35.60,19.12}&
\multicolumn{1}{c|}{21.71,9.55}&
 \multicolumn{1}{c}{\textbf{8.54,3.23}}\\
\hline
LOOP2&
 \multicolumn{1}{c|}{31.94,16.42}&
 \multicolumn{1}{c|}{23.06,12.27}&
\multicolumn{1}{c}{\textbf{8.57,3.94}}\\
\hline
Average&
\multicolumn{1}{c|}{33.77, 17.77}&
\multicolumn{1}{c|}{22.38, 10.91}&
\multicolumn{1}{c}{\textbf{8.55, 3.58}}\\
\hline
\end{tabular}}
\vspace{-2mm} 
\end{table}

\begin{table}[htp]
    \centering
\caption{\textbf{Training and testing Sequences of Oxford RobotCar.}} 
\begin{tabular}{c|c|c|c} 
\hline
Sequence	 &Time	 &Tag	 &Mode\\
\hline
-	&2014-06-26-08-53-56	&overcast	&Training	\\
-	&2014-06-26-09-24-58	&overcast	&Training	\\
LOOP1 	&2014-06-23-15-41-25	&sunny	&Testing\\
LOOP2	&2014-06-23-15-36-04	&sunny	&Testing\\
\hline
\end{tabular}\vspace{-6mm}
\end{table}

\subsubsection{Visualization Analysis}

To investigate directional attention in camera localization, we analyzed our model's and AtLoc's saliency maps on the RobotCar dataset during straight and turning (Figure~\ref{saliency}). When driving straight, directional attention makes NeuroLoc learn stable geometric elastic object structures (e.g., building-body intersections, trees, and skyline in Figure~\ref{saliencyB} (top)). In contrast, AtLoc learns few static environmental features (e.g., roads in Figure~\ref{saliencyA} (top)). This shows that our model has better attention localization in straight-ahead scenarios and enhanced global-localization robustness. When turning, directional attention enables NeuroLoc to generate a response mechanism like head direction cells in attention localization, learning explicit feature-direction correspondences (e.g., in Figure~\ref{saliencyB} (bottom), attention focuses on building edges corresponding to their true orientation; in Figure~\ref{saliencyB} (top), attention is more dispersed among roads, trees, and fences). This enhances the rotation-prediction accuracy in turning scenarios.
\subsection{Ablation Study}
We conducted ablation experiments on the Oxford RoboCar dataset. The ablation model settings are as follows: 1) We will remove the Hebbian storage module and pose regression module from NeuroLoc and use an attention network and fully connected layer as the NeuroLoc-Base. 2) We will add the Hebbian storage module to NeuroLoc Base and use it as the Neuro-Hebbian. 3) NeuroLoc is our complete model. Table~\ref{ablation} shows that by sequentially adding brain inspired modules to the attention-based pose regression model, there is a significant improvement in position and rotation prediction performance.

\section{CONCLUSIONS}
%
Camera localization is a challenging task in computer vision due to scene dynamics and the high variability of environment appearance. The proposed NeuroLoc is inspired by the navigation cells. In NeuroLoc, the Hebbian storage module reduces scene ambiguity, and directional attention can guide the network to learn robust geometric features, which enables our method to achieve state-of-the-art performance. Further work includes investigating whether other mechanisms of the navigation cells can improve the robustness and adaptability of camera localization.













\bibliographystyle{IEEEtran}
\bibliography{References}

\end{document}